\documentclass[runningheads]{./styles/llncs}
\usepackage{graphicx}
\usepackage{url}
\usepackage{hyperref, booktabs}
\usepackage{subcaption, multirow}
\usepackage{chngpage}
\usepackage{amsfonts, amsmath, amssymb, bm, bbm, mathtools}
\usepackage{float, colortbl, tabularx, longtable, multirow, environ, wrapfig, textcomp, nicefrac}
\usepackage{algorithm,algorithmic}
\usepackage[table,dvipsnames]{xcolor}
\usepackage{mdwlist}

\def \e  {\epsilon}
\def \pa {\text{pa}}

\begin{document}
\title{Harmonization with Flow-based\\ Causal Inference}
\titlerunning{Harmonization with Flow-based Causal Inference}
%
\author{Rongguang Wang\inst{1,2} \and
Pratik Chaudhari\inst{1,3} \and
Christos Davatzikos\inst{1,2,4}}
%
\authorrunning{R. Wang et al.}
%
\institute{
Department of Electrical and Systems Engineering, University of Pennsylvania
\and
Center for Biomedical Image Computing and Analytics (CBICA)
\and
General Robotics, Automation, Sensing and Perception Laboratory (GRASP)
\and
Department of Radiology, Perelman School of Medicine, University of Pennsylvania\\
\email{\{rgw@seas, pratikac@seas, Christos.Davatzikos@pennmedicine\}.upenn.edu}
}
\maketitle              

\begin{abstract}


Heterogeneity in medical data, e.g., from data collected at different sites and with different protocols in a clinical study, is a fundamental hurdle for accurate prediction using machine learning models, as such models often fail to generalize well. This paper leverages a recently proposed normalizing-flow-based method to perform counterfactual inference upon a structural causal model (SCM), in order to achieve harmonization of such data. A causal model is used to model observed effects (brain magnetic resonance imaging data) that result from known confounders (site, gender and age) and exogenous noise variables. Our formulation exploits the bijection induced by flow for the purpose of harmonization. We  infer the posterior of exogenous variables, intervene on observations, and draw samples from the resultant SCM to obtain counterfactuals. This approach is evaluated extensively on multiple, large, real-world medical datasets and displayed better cross-domain generalization compared to state-of-the-art algorithms. Further experiments that evaluate the quality of confounder-independent data generated by our model using regression and classification tasks are provided.

\end{abstract}

\section{Introduction}

Deep learning models have shown great promise in medical imaging diagnostics~\cite{esteva2017dermatologist} and predictive modeling with applications ranging from segmentation tasks~\cite{menze2014multimodal} to more complex decision-support functions for phenotyping brain diseases and personalized prognosis. However deep learning models tend to have poor reproducibility across hospitals, scanners, and patient cohorts; these high-dimensional models tend to overfit to specific datasets and generalize poorly across training data~\cite{davatzikos2019machine}. One potential solution to the above problem is to train on very large and diverse databases but this can be prohibitive, because data may change frequently (e.g., new imaging devices are introduced) and gathering training labels for medical images is expensive. More importantly, even if it were possible to train a model on data that covers all possible variations across images, such a model would almost certainly sacrifice accuracy in favor of generalization---it would rely on coarse imaging features that are stable across, say imaging devices and patient populations, and might fail to capture more subtle and informative detail. Methods that can tackle heterogeneity in medical data without sacrificing predictive accuracy are needed, including methods for ``data harmonization'',  which would allow training a classifier on, say data from one site, and obtaining similar predictive accuracy on data from another site.

\paragraph{Contributions}
We build upon a recently proposed framework~\cite{pawlowski2020deep} for causal inference, by modeling brain imaging data and clinical variables via a causal graph and focus on how causes (site, gender and age) result in the effects, namely imaging measurements (herein we use region of interest (ROI) volumes obtained by preprocessing brain MRI data). This framework uses a normalizing flow parameterized by deep networks to learn the structural assignments in a causal graph. We demonstrate how harmonization of data can be performed efficiently using counterfactual inference on such a flow-based causal model. 
Given a dataset pertaining to one site (source), we perform a counterfactual query to synthesize the dataset, as if it were from another site (target). Essentially, this amounts to the counterfactual question ``what would the scans look like if they had been acquired from the same site''. We demonstrate results of such harmonization on regression (age prediction) and classification (predicting Alzheimer's disease) tasks using several large-scale brain imaging datasets. We demonstrate substantial improvement over competitive baselines on these tasks. 

\section{Related Work}
\label{sec:related_works}

A wide variety of recent advances have been made to remove undesired confounders for imaging data, e.g., pertaining to sites or scanners~\cite{johnson2007adjusting,pomponio2020harmonization,moyer2018invariant,moyer2020scanner,robinson2020image,bashyam2020medical}. Methods like ComBat~\cite{johnson2007adjusting,pomponio2020harmonization}, based on parametric empirical Bayes~\cite{morris1983parametric}, produce site-removed image features by performing location (mean) and scale (variance) adjustments to the data. A linear model estimates location and scale differences in images features across sites while preserving confounders such as sex and age.
In this approach, other unknown variations such as race and disease are removed together with the site variable, which might lead to inaccurate predictions for disease diagnosis. Generative deep learning models such as variational autoencoders (VAEs)~\cite{kingma2013auto} and generative adversarial networks (GANs)~\cite{goodfellow2014generative} have been used in many works~\cite{moyer2018invariant,moyer2020scanner,robinson2020image,bashyam2020medical}.
These methods typically minimize the mutual information between the site variable and image embedding in the latent space, and learn a site-disentangled representation which can be used to reconstruct images from a different site. Unsupervised image-to-image translation has been used to map scans either between two sites~\cite{robinson2020image} or to a reference domain~\cite{bashyam2020medical} using models like CycleGAN~\cite{zhu2017unpaired}. Generative models are however challenging to use in practice: VAEs typically suffer from blurry reconstructions while GANs can suffer from mode collapse and convergence issues. These issues are exacerbated for 3D images. In this paper, we focus on regions of interest (ROI) features. We extend the deep structural causal model of~\cite{pawlowski2020deep} which enables tractable counterfactual inference from single-site healthy MR images to multi-center pathology-associated scans for data harmonization. Besides qualitative examination of the counterfactuals performed in~\cite{pawlowski2020deep}, we provide extensive quantitative evaluations and compare it with state-of-the-art harmonization baselines.

\section{Method}


Our method builds upon the causal inference mechanism proposed by Judea Pearl~\cite{pearl2009causality} and method of Pawlowski et al.~\cite{pawlowski2020deep} that allows performing counterfactual queries upon causal models parameterized by deep networks. We first introduce preliminaries of our method, namely, structural causal models, counterfactual inference, and normalizing flows, and then describe the proposed harmonization algorithm.

\subsection{Building blocks}


\paragraph{Structural Causal Models (SCMs).} are analogues of directed probabilistic graphical models for causal inference~\cite{peters2017elements,scholkopf2019causality}. Parent-child relationships in an SCM denote the effect (child) of direct causes (parents) while they only denote conditional independencies in a graphical model. Consider a collection of random variables $x = (x_1,\ldots,x_m)$, an SCM given by $M = (S, P_\e)$ consists of a collection $S = (f_1,\ldots,f_m)$ of assignments $x_k = f_k(\e_k; \pa_k)$ where $\pa_k$ denotes the set of parents (direct causes) of $x_k$ and noise variables $\e_k$ are unknown and unmodeled sources of variation for $x_k$. Each variable $x_k$ is independent of its non-effects given its direct causes (known as the causal Markov condition), we can write the joint distribution of an SCM as $P_M(x) = \prod_{k=1}^m P(x_k\ |\ \pa_k)$; each conditional distribution here is determined by the corresponding structural assignment $f_k$ and noise distribution~\cite{pearl2009causality}. Exogenous noise variables are assumed to have a joint distribution $P_\e = \prod_{k=1}^m P(\e_i)$, this will be useful in the sequel.

\paragraph{Counterfactual Inference.}

Given a SCM, a counterfactual query is formulated as a three-step process: abduction, action, and prediction~\cite{pearl2009causality,peters2017elements,pearl2009causal}. First, we predict exogenous noise $\e$ based on observations to get the posterior $P_M(\e\ |\ x) = \prod_{k=1}^m P_M(\e_k\ |\ x_k, \pa_k)$. Then comes intervention denoted by $\text{do}(\tilde{x}_k)$, where we replace structural assignments of variable $x_k$. Intervention makes the effect $x_k$ independent of both its causes $\pa_k$ and noise $\e_k$ and this results in a modified SCM $\tilde{M} = M_{\text{do}(\tilde{x})} \equiv (\tilde{S}, P_M(\e\ |\ x))$. Note that the noise distribution has also been modified, it is now the posterior $P_M(\e\ |\ x)$ obtained in the abduction step. The third step, namely prediction involves predicting counterfactuals by sampling from the distribution $P_{\tilde{M}}(x)$ entailed by the modified SCM.

\begin{figure}[!t]
\centering
\includegraphics[width=0.35\textwidth]{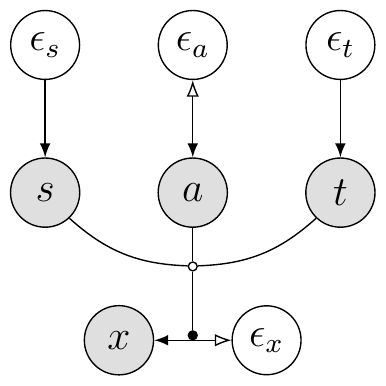}
\caption{Causal graph of the structural causal model for brain imaging. Data consists of brain ROIs ($x$), sex ($s$), age ($a$), imaging site ($t$), and their respective exogenous variables ($\epsilon_x$, $\epsilon_s$, $\epsilon_a$, and $\epsilon_t$). Bidirectional arrows indicate invertible normalizing flow models and the black dot shows that the flow model associated with $x$ is conditioned on the direct causes (parents) $s$, $a$, and $t$; this follows the notation introduced in~\cite{pawlowski2020deep}.
We are interested in answering counterfactual questions of the form ``what would the scans look like if they had been acquired from the same site''. We first train the flow-based SCM $M_\theta$ on the observed data.
We then infer the posterior of exogenous variables $\epsilon_x$ and $\epsilon_a$ with the invertible structural assignments (abduction step). We can now intervene upon the site by replacing site variable $t$ with a specific value $\tau$, this is denoted by $\text{do}(t=\tau)$. We sample from the modified flow-based SCM $M_{\text{do}(t=\tau)}$ to obtain counterfactual queries.
}
\label{fig:model}
\end{figure}

\paragraph{Learning a normalizing flow-based SCM.}
Given the structure of the SCM, learning the model involves learning the structure assignments $S$ from data. Following~\cite{pawlowski2020deep}, we next do so using normalizing flows parameterized by deep networks. Normalizing flows model a complex probability density as the result of a transformation applied to some simple probability density~\cite{papamakarios2019normalizing,papamakarios2017masked,dolatabadi2020invertible,durkan2019neural}; these transformations are learned using samples from the target. Given observed variables $x$ and base density $\e \sim p(\e)$, this involves finding an invertible and differentiable transformation $x = f(\e)$. The density of $x$ is given by $p(x) = p(\e)\ |\text{det} \nabla f(\e)|^{-1}$ where $\e = f^{-1}(x)$ and $\nabla f(\e)$ is the Jacobian of the flow $f: \e \mapsto x$. The density $p(\e)$ is typically chosen to be a Gaussian. Given a dataset $D = \{x^i \sim p(x) \}_{i=1}^n$ with $n$ samples, a $\theta$-parameterized normalizing flow $f_\theta$ can fitted using a maximum-likelihood objective to obtain
\[
    \theta^* = \text{argmax}\  \frac{1}{n} \sum_{i=1}^n \log p(\e^i)\  - \log | \text{det} \nabla f_\theta(\e^i)|.
\]
Here $e^i = f_\theta^{-1}(x^i)$. Parameterizing the normalizing flow using a deep network leads to powerful density estimation methods. This approach can be easily extended to conditional densities of the form $p(x_k \mid \pa_k)$ in our SCM.

\subsection{Harmonization using counterfactual inference in a flow-based SCM}

Given the structure of a SCM, we fit conditional flows $f_{\theta_k}: \e_k \mapsto x_k$ that map exogenous noise to effect $x_k$ given parents $\pa_k$ for all nodes in the SCM. We will denote the combined flow for all nodes in the SCM as $f_\theta$ which maps noise $\e^i = (\e^i_1,\ldots,\e^i_m)$ to observations $x^i = (x^i_1,\ldots,x^i_m)$ in the dataset; the corresponding SCM is denoted by $M_\theta$. Focus on a particular datum $x^i$ in the dataset. The abduction step simply computes $\e^i = f_\theta^{-1}(x^i)$. Formally this corresponds to computing the posterior distribution $P_{M_\theta}(\e \mid x^i)$. Intervention uses the fact that the flow models a conditional distribution and replaces (intervenes) the value of a particular variable, say $x^i_k \leftarrow \tilde{x}_k^i$; this corresponds to the operation $\text{do}(\tilde{x}_k)$. The variable $x_k$ is decoupled from its parents and exogenous noise which corresponds to a modified structural assignment $\tilde{f}_{\theta_k}$ and results in a new SCM $\tilde{M}_\theta$. We can now run the same flow $\tilde{f}_\theta$ forwards using samples $\e^i$ from the abduction step to get samples from $P_{\tilde{M}_\theta}(x)$ which are the counterfactuals. Fig.~\ref{fig:model} shows an example SCM for brain imaging data and shows we perform counterfactual queries to remove site effects.


\section{Experimental Results}

\subsection{Setup}

\paragraph{Datasets.}
We use 6,921 3D T1-weighted brain magnetic resonance imaging (MRI) scans acquired from multiple scanners or sites in Alzheimer’s Disease Neuroimaging Initiative (ADNI)~\cite{jack2008alzheimer} and the iSTAGING consortium~\cite{habes2021brain} which consists of Baltimore Longitudinal Study of Aging (BLSA)~\cite{resnick2003longitudinal,armstrong2019predictors}, Study of Health in Pomerania (SHIP)~\cite{hegenscheid2009whole} and the UK Biobank (UKBB)~\cite{sudlow2015uk}. Detailed demographic information of the datasets is provided in the Appendix. We first perform a sequence of preprocessing steps on these images, including bias-filed correction~\cite{tustison2010n4itk}, brain tissue extraction via skull-stripping~\cite{doshi2013multi}, and multi-atlas segmentation~\cite{doshi2016muse}. Each scan is then segmented into 145 anatomical regions of interests (ROIs) spanning the entire brain, and finally volumes of the ROIs are taken as the features. We first perform age prediction task using data from the iSTAGING consortium for participants between ages 21--93 years. We then demonstrate our method for classification of Alzheimer’s disease (AD) using the ADNI dataset where the diagnosis groups are cognitive normal (CN) and AD; this is a more challenging problem than age prediction.

\paragraph{Implementation.}
We implement flow-based SCM with three different flows (affine, linear and quadratic autoregressive splines~\cite{dolatabadi2020invertible,durkan2019neural}) using PyTorch~\cite{paszke2019pytorch} and Pyro~\cite{bingham2019pyro}. We use a categorical distribution for sex and site, and real-valued normalizing flow for other structural assignments. A linear flow and a conditional flow (conditioned on activations of a fully-connected network that takes age, sex and scanner ID as input) are used as structural assignments for age and ROI features respectively. The density of exogenous noise is standard Gaussian. For training, we use Adam~\cite{kingma2014adam} with batch-size of 64, initial learning rate $3 \times 10^{-4}$ and weight decay $10^{-4}$. We use a staircase learning rate schedule with decay milestones at $50\%$ and $75\%$ of the training duration. All models are trained for at least 100 epochs. Implementation details for the SCM and the classifier, and the best validation log-likelihood for each model are shown in the Appendix.

\paragraph{Baselines.}

We compare with a number of state-of-the-art algorithms: invariant risk minimization (IRM)~\cite{arjovsky2019invariant}, ComBat~\cite{johnson2007adjusting,pomponio2020harmonization}, ComBat$++$~\cite{wachinger2021detect}, and CovBat~\cite{chen2020removal} on age regression and Alzheimer’s disease classification. IRM learns an invariant representation that the optimal predictor using this representation is simultaneously optimal in all environments. We implement IRM and ComBat algorithms with publicly available code from the original authors. We also show results obtained by training directly on the target data which acts as upper-bound on the accuracy of our harmonization.

\subsection{Evaluation of the learned flow-based SCM}

\begin{figure*}[!t]
\centering
\includegraphics[width=1.\textwidth]{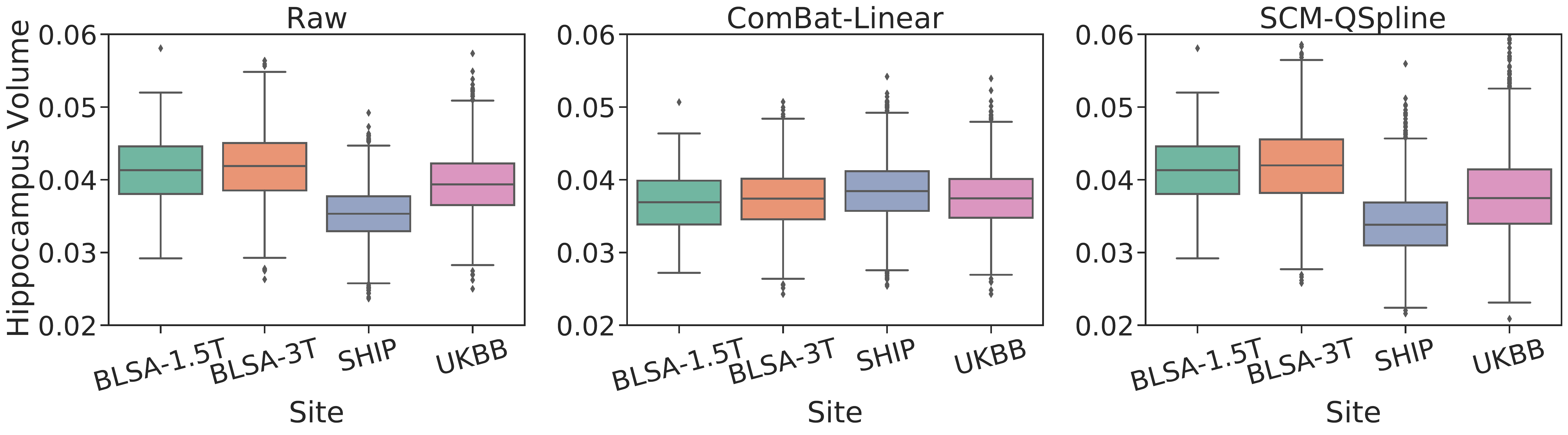}
\caption{Comparison of normalized feature (hippocampus volume) distributions for various sites in the iSTAGING consortium data before (raw) and after harmonization using ComBat-Linear and our SCM-QSpline.
We observe that ComBat aligns inter-site feature distributions by preserving sex and age effects and removes all other unknown confounders by treating them as site effects. In contrast, the distribution of hippocampus volume is unchanged in our proposed method which takes both known confounders (sex, age, and site) and unknown confounders (exogenous noises) into consideration for harmonization. ComBat removing these useful confounders is detrimental to accuracy (see Table~\ref{tab:adni_result}).
}
\label{fig:feature_distribution}
\end{figure*}

We explore three normalizing flow models: affine, linear autoregressive spline~\cite{dolatabadi2020invertible}, and quadratic autoregressive spline~\cite{durkan2019neural}. Implementation details and their log-likelihood are in the Appendix. For both iSTAGING and ADNI datasets, the log-likelihood improves consistently with the model's expressive power.
Spline-based autoregressive flow models (17.22 log-likelihood for linear-spline and 17.24 for quadratic-spline) are better for density estimation than an affine flow model (1.88 log-likelihood). A quadratic spline-based model obtains slightly higher log-likelihood than the linear-spline model on the iSTAGING dataset.

We next show the feature (hippocampus volume) distributions of raw data, ComBat~\cite{johnson2007adjusting,pomponio2020harmonization} transformed data and the data generated from the flow-based SCM in Fig.~\ref{fig:feature_distribution}. We find that the feature distributions of ComBat are not consistent with those of raw data; ComBat transformed feature distributions show similar means (all shifted to an average value across sites) which arises from removing site-dependent location and scale effects. The third panel shows data generated from counterfactual queries with the flow-based SCM fitted on BLSA-3T. Feature distributions for the SCM-data are similar to those of the raw data. This can be attributed to the fact that our method preserves the unknown confounders (subject-specific information due to biological variability, such as race, gene, and pathology AD/CN) by capturing them as exogenous noise in the SCM.

\subsection{Age prediction}

\begin{table}[!t]
\caption{Mean average error of age prediction for data from the iSTAGING consortium. All experiments were repeated 5 times in cross-validation fashion, and the average performance is reported with the standard errors in the brackets. TarOnly indicates validation MAEs directly trained on each target sites. The hypothesis that our proposed methods achieve a better accuracy than the baselines can be accepted with $p$-values between 0.06--0.41. This task can be interpreted as a sanity check for our method.}
\label{tab:istaging_result}
\begin{center}
\begin{scriptsize}
\begingroup
\setlength{\tabcolsep}{2.1pt}
\resizebox{\columnwidth}{!}{
\begin{tabular}{l cccc |cccc|ccc }
\toprule
\multirow{2}{*}{ } & \multirow{2}{*}{Study}  & \multirow{2}{*}{TarOnly} & \multirow{2}{*}{SrcOnly}  & \multirow{2}{*}{IRM}  & \multicolumn{4}{c}{ComBat} & \multicolumn{3}{|c}{Flow-based SCM (ours)} \\
& & & & & Linear & GAM & ComBat++ & CovBat & Affine & L-Spline & Q-Spline\\
\midrule
\multirow{2}{*}{Source} & \multirow{2}{*}{BLSA-3T} & \multirow{2}{*}{-} & 11.74  & 11.76
& 11.72  & 11.74 & 11.73 & 11.74
& 11.74  & 11.74  & 11.65 \\
& & & (0.35) & (0.35) & (0.62) & (0.61) & (0.62) & (0.62) & (0.61) & (0.61) & (0.62) \\
\multirow{2}{*}{Target} & \multirow{2}{*}{BLSA-1.5T} & 6.77  & 7.21  & 7.16
& 7.14  & 7.01 & 7.00 & 7.03
& 7.01  & 7.00  & 6.92 \\
& & (0.82) & (0.91) & (0.87) & (0.99) & (0.99) & (1.04) & (1.08) & (1.01) & (1.04) & (1.09) \\
\multirow{2}{*}{Target} & \multirow{2}{*}{UKBB} & 6.14  & 7.27  & 7.18
& 6.62  & 6.70 & 6.71 & 6.75
& 6.72  & 6.75  & 6.44 \\
& & (0.16) & (0.70) & (0.58) & (0.46) & (0.46) & (0.47) & (0.49) & (0.46) & (0.47) & (0.28) \\
\multirow{2}{*}{Target} & \multirow{2}{*}{SHIP} & 11.36  & 17.14  & 17.05
& 15.95  & 16.17 & 16.21 & 16.22
& 16.20  & 16.25  & 15.68 \\
& & (0.31) & (0.62) & (0.46) & (0.61) & (0.59) & (0.47) & (0.65) & (0.59) & (0.63) & (0.80) \\
\bottomrule
\end{tabular}
}
\endgroup
\end{scriptsize}
\end{center}
\end{table}

In Table~\ref{tab:istaging_result}, we compare the mean average error (MAE) of age prediction for a regression trained on raw data, site-removed data generated by ComBat~\cite{johnson2007adjusting,pomponio2020harmonization} and its variants~\cite{wachinger2021detect,chen2020removal}, IRM~\cite{arjovsky2019invariant} and counterfactuals generated by our flow-based SCM. All models are trained on BLSA-3T (source site) and then tested on BLSA-1.5T, UKBB, and SHIP separately. We find that model (SrcOnly) trained on the source site with raw data cannot generalize to data from the target site. Models trained with site-removed data generated by ComBat generalize much better compared ones trained on raw data (SrcOnly), whereas IRM shows marginal improvement compared to SrcOnly.
All variants (affine, linear-spline, quadratic-spline) of flow-based SCM show substantially smaller MAE; quadratic-spline SCM outperforms the other methods on all target sites.

\subsection{Classification of Alzheimer's Disease}

\begin{table}[!t]
\caption{AD classification accuracy (\%) comparison on the ADNI dataset and standard deviation (in brackets) across 5-fold cross-validation. TarOnly indicates validation accuracies for training directly on each target site. The hypothesis that our proposed method (Q-Spline) achieves a better accuracy than baselines can be accepted with $p$-values less than $10^{-5}$.}
\label{tab:adni_result}
\begin{center}
\begin{scriptsize}
\begingroup
\setlength{\tabcolsep}{2.1pt}
\resizebox{\columnwidth}{!}{
\begin{tabular}{l cccc |cccc|ccc }
\toprule
\multirow{2}{*}{ } & \multirow{2}{*}{Study} & \multirow{2}{*}{TarOnly} & \multirow{2}{*}{SrcOnly} & \multirow{2}{*}{IRM} & \multicolumn{4}{c}{ComBat} & \multicolumn{3}{|c}{Flow-based SCM (ours)} \\
& & & & & Linear & GAM & ComBat++ & CovBat & Affine & L-Spline & Q-Spline \\
\midrule
\multirow{2}{*}{Source} & \multirow{2}{*}{ADNI-1} & \multirow{2}{*}{-} & 76.1  & 76.2
& 75.1  & 75.1 & 65.1 & 74.4
& 76.1  & 75.3  & 75.4  \\
& & & (1.54) & (2.46) & (1.37) & (1.23) & (6.29) & (2.29) & (1.92) & (1.76) & (2.45) \\
\multirow{2}{*}{Target} & \multirow{2}{*}{ADNI-2} & 75.8  & 71.9  & 73.0
& 71.4  & 72.1 & 56.2 & 67.4
& 73.4  & 72.6  & 73.7 \\
& & (3.46) & (4.88) & (4.85) & (4.30) & (2.83) & (9.29) & (5.06) & (3.52) & (3.48) & (4.13) \\
\midrule
\multirow{2}{*}{Source} & \multirow{2}{*}{ADNI-2} & \multirow{2}{*}{-} & 75.8  & 76.3
& 77.5  & 77.0 & 67.8 & 77.9
& 78.7  & 78.2  & 77.5  \\
& & & (3.46) & (2.35) & (2.30) & (2.74) & (9.42) & (2.47) & (1.32) & (2.80) & (1.76) \\
\multirow{2}{*}{Target} & \multirow{2}{*}{ADNI-1} & 76.1  & 70.4  & 72.0
& 71.1  & 70.1 & 58.0 & 69.1
& 71.4  & 71.8  & 73.3 \\
& & (1.54) & (8.80) & (2.16) & (4.07) & (5.67) & (6.28) & (5.82) & (2.41) & (5.76) & (3.04) \\
\bottomrule
\end{tabular}
}
\endgroup
\end{scriptsize}
\end{center}
\end{table}

In Table~\ref{tab:adni_result}, we show the accuracy of a classifier trained on raw data, Combat-harmonized data and SCM-generated counterfactuals for the ADNI dataset; this is a binary classification task with classes being CN (cognitive normal) and AD. All classifiers are trained on source sites (ADNI-1 or ADNI-2) and evaluated on target sites (ADNI-2 or ADNI-1 respectively). The classifier works poorly on the target site without any harmonization (SrcOnly).
ComBat-based methods show a smaller gap between the accuracy on the source and target site; IRM improves upon this gap considerably. Harmonization using our flow-based SCM, in particular the Q-Spline variant, typically achieves higher accuracies on the target site compared to these methods.


\section{Conclusion}

This paper tackles harmonization of data from different sources using a method inspired from the literature on causal inference~\cite{pawlowski2020deep}.
The main idea is to explicitly model the causal relationship of known confounders such as sex, age, and site, and ROI features in a SCM that uses normalizing flows to model probability distributions. Counterfactual inference can be performed upon such a model to sample harmonized data by intervening upon these variables. We demonstrated experimental results on two tasks, age regression and Alzheimer's Disease classification, on a wide range of real-world datasets. We showed that our method compares favorably to state-of-the-art algorithms such as IRM and ComBat. Future directions for this work include causal graph identification and causal mediation.

\section*{Acknowledgements}
We thank Ben Glocker, Nick Pawlowski and Daniel C. Castro for suggestions. 
This work was supported by the National Institute on Aging (grant numbers RF1AG054409 and U01AG068057) and the National Institute of Mental Health (grant number R01MH112070).
Pratik Chaudhari would like to acknowledge the support of the Amazon Web Services Machine Learning Research Award.

%
%

\bibliographystyle{./styles/splncs04}
\bibliography{refs}

\clearpage
\section*{Appendix}

\begin{table}[!h]
\caption{Summary of participant demographics in iSTAGING consortium. Age is described in format: mean $\pm$ std [min, max]. F and M in gender represent female and male separately. Field indicates the magnetic strength of the MRI scanners.}
\label{tab:istaging_data}
\begin{center}
\begingroup
\setlength{\tabcolsep}{5pt}
\begin{tabular}{l cccc }
\toprule
Study & Subject & Age & Gender (F/M) & Field  \\
\midrule
BLSA-1.5T & 157 & 69.1 $\pm$ 8.5 [48.0, 85.0]  & 66 / 91 & 1.5T \\
BLSA-3T & 960 & 65.0 $\pm$ 14.7 [22.0, 93.0] & 525 / 435 & 3T \\
UKBB & 2202 & 62.8 $\pm$ 7.3 [45.0, 79.0] & 1189 / 1013 & 3T \\
SHIP & 2739 & 52.6 $\pm$ 13.7 [21.2, 90.4] & 1491 / 1248 & 1.5T \\
\bottomrule
\end{tabular}
\endgroup
\end{center}
\end{table}

\begin{table}[!h]
\caption{Summary of participant demographics in ADNI dataset. Age is described in format: mean $\pm$ std [min, max]. F and M in gender represent female and male separately. Field indicates the magnetic strength of the MRI scanners.}
\label{tab:adni_data}
\begin{center}
\begingroup
\setlength{\tabcolsep}{5pt}
\begin{tabular}{l cccccc }
\toprule
Study & Subject & CN & AD & Age & Gender (F/M) & Field  \\
\midrule
ADNI-1 & 422 & 229 & 193 & 75.5 $\pm$ 6.2 [55.0, 90.9]  & 201 / 221 & 1.5T \\
ADNI-2/GO & 441 & 294 & 147 & 73.4 $\pm$ 6.8 [55.4, 90.3] & 221 / 220 & 3T \\
\bottomrule
\end{tabular}
\endgroup
\end{center}
\end{table}

\begin{table}[!h]
\caption{Multi-layer perceptron (MLP) network implementation details. The network is used for age regression and AD classification tasks. The output size $k$ of the final layer is depends on the task.}
\label{tab:arch_c}
\begin{center}
\begin{small}
\begingroup
\setlength{\tabcolsep}{5pt}
\begin{tabular}{l ccc }
\toprule
Layer & Input Size & LeakyReLU $\alpha$  & Output Size \\
\midrule
Linear + LeakyReLU & 145 & 0.1 & 72 \\
Linear + LeakyReLU & 72 & 0.1 & 36 \\
Linear & 36 & - & k \\
\bottomrule
\end{tabular}
\endgroup
\end{small}
\end{center}
\end{table}

\begin{table}[!h]
\caption{Flow-based SCM implementation details. We directly learn the binary probability of sex $s$ and categorical probability of site $t$. $p_\theta^S$ and $p_\theta^T$ are the learnable mass functions of the categorical distribution for variables sex $s$ and site $t$, and $K$ is the number of site $t$. The modules indicated with $\theta$ are parameterized using neural networks. We constrain age $a$ variable with lower bound (exponential transform) and rescale it with fixed affine transform for normalization. $\text{Spline}_\theta$ transformation refers to the linear neural spline flows~\cite{dolatabadi2020invertible}.
The $\text{ConditionalTransform}_\theta(\cdot)$ can be conditional affine or conditional spline transform, which reparameterizes the noise distribution into another Gaussian distribution. We use linear~\cite{dolatabadi2020invertible} and quadratic~\cite{durkan2019neural} autoregressive neural spline flows for the conditional spline transform, which are more expressive compared to the affine flows. The transformation parameters of the $\text{ConditionalTransform}_\theta(\cdot)$ are predicted by a context neural network taking $\cdot$ as input. The context networks are implemented as fully-connected networks for affine and spline flows.}
\begin{center}
\begin{small}
\begingroup
\setlength{\tabcolsep}{5pt}
\begin{tabular}{l l}
\toprule
Observations & Exogenous noise  \\
\midrule
$s:= \e_S$ & $\e_S\sim \text{Ber}(p_\theta^S)$  \\
$a:= f_A(\e_A)=(\text{Spline}_\theta \circ \text{Affine} \circ \text{Exp})(\e_A)$ & $\e_A\sim \mathcal{N}(0, 1)$  \\
$t:= \e_T$ & $\e_T\sim \text{Cat}(K, p_\theta^T)$  \\
$x:= f_X(\e_X; s, a, t)=(\text{ConditionalTransform}_\theta([s, a, t]))(\e_X)$ & $\e_X\sim \mathcal{N}(0, 1)$ \\
\bottomrule
\end{tabular}
\endgroup
\end{small}
\end{center}
\end{table}

\begin{table}[!h]
\caption{Comparison of associative abilities of different type of flows on iSTAGING consortium and ADNI dataset. We observe that spline flows achieved higher log-likelihood compared to that of affine flow for both datasets. This indicates that a flow with higher expressive power helps for density estimation.}
\label{tab:density_result}
\begin{center}
\begin{small}
\begingroup
\setlength{\tabcolsep}{5pt}
\begin{tabular}{l cc }
\toprule
Study & Model & Log-likelihood \\
\midrule
\multirow{3}{*}{iSTAGING} & Affine & 1.8817 \\
& Linear Spline & 17.2204 \\
& Quadratic Spline & 17.2397 \\
\midrule
\multirow{3}{*}{ADNI} & Affine & 1.8963 \\
& Linear Spline & 15.2715 \\
& Quadratic Spline & 15.2055 \\
\bottomrule
\end{tabular}
\endgroup
\end{small}
\end{center}
\end{table}


\begin{figure}[!h]
\centering
\begin{subfigure}[t]{\linewidth}
\centering
\includegraphics[width=\linewidth]{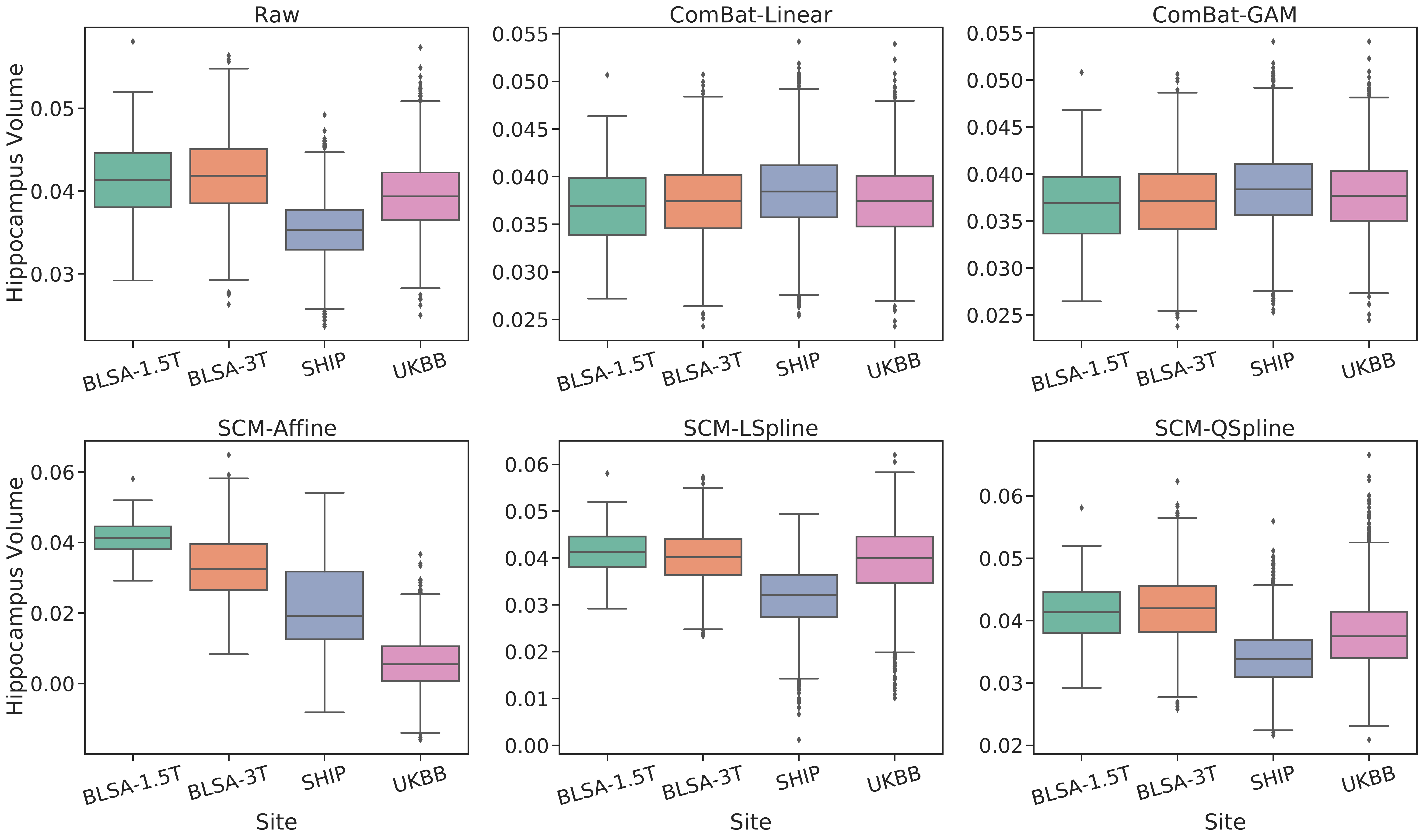}
\caption{Hippocampus (Right)}
\end{subfigure}

\begin{subfigure}[t]{\linewidth}
\centering
\includegraphics[width=\linewidth]{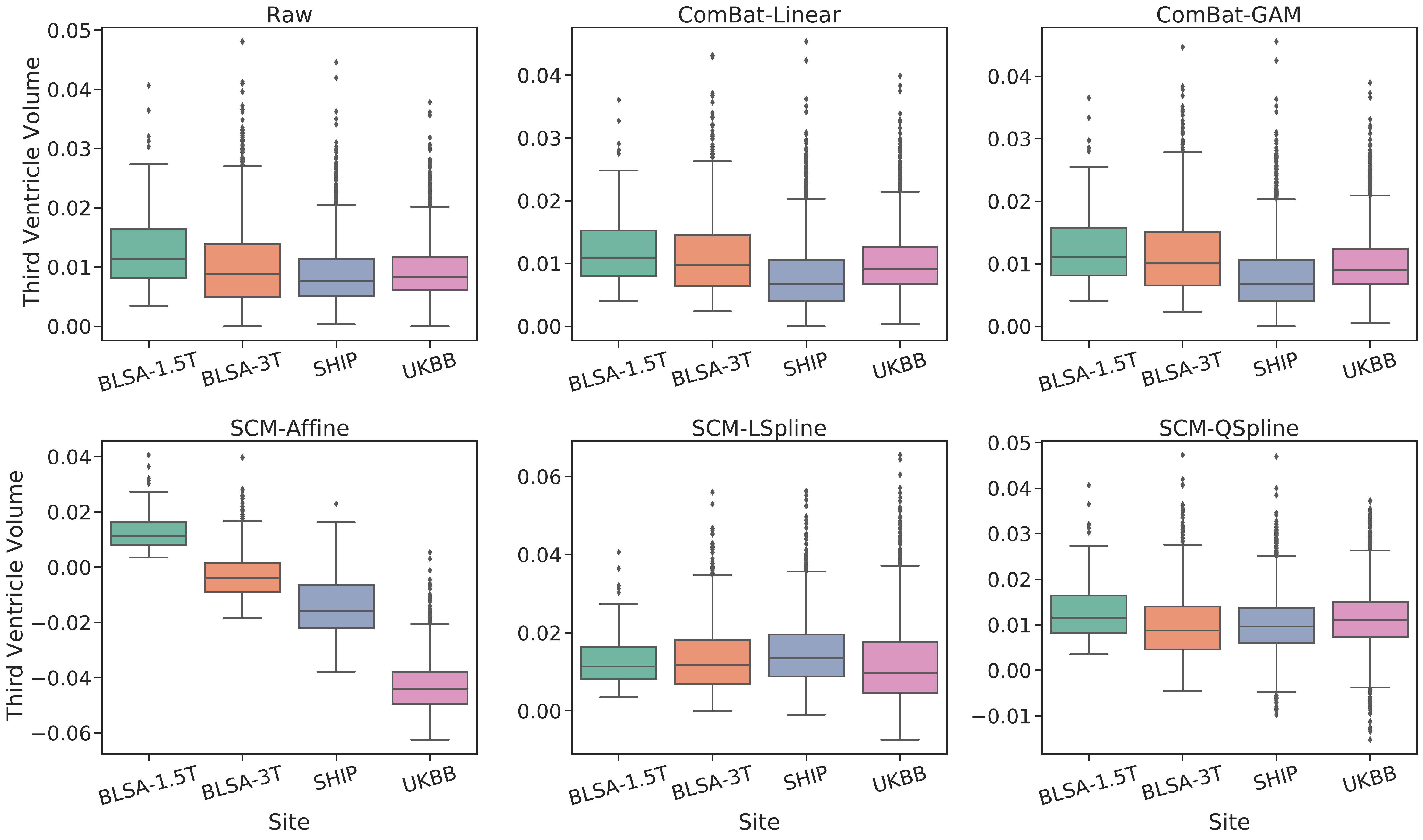}
\caption{Third Ventricle (Right)}
\end{subfigure}
\caption{Comparison of normalized feature distributions cross-site in iSTAGING consortium before and after apply the ComBat methods (ComBat-Linear and ComBat-GAM) and the proposed methods (SCM-Affine, SCM-LSpline, and SCM-QSpline).
The distributions of the features harmonized by ComBat methods are aligned cross-site, whereas those harmonized by our proposed method (Q-Spline) are unchanged compared to the raw features.
We preserve the unknown cofounders (subject-specific information due to biological variability, such as race, gene, and pathology AD/CN) instead of removing them as site-effects, which is beneficial for downstream analysis, such as AD diagnosis.
}
\label{fig:third_ventricle}
\end{figure}

\begin{figure}[!h]
\centering
\begin{subfigure}[t]{\linewidth}
\centering
\includegraphics[width=\linewidth]{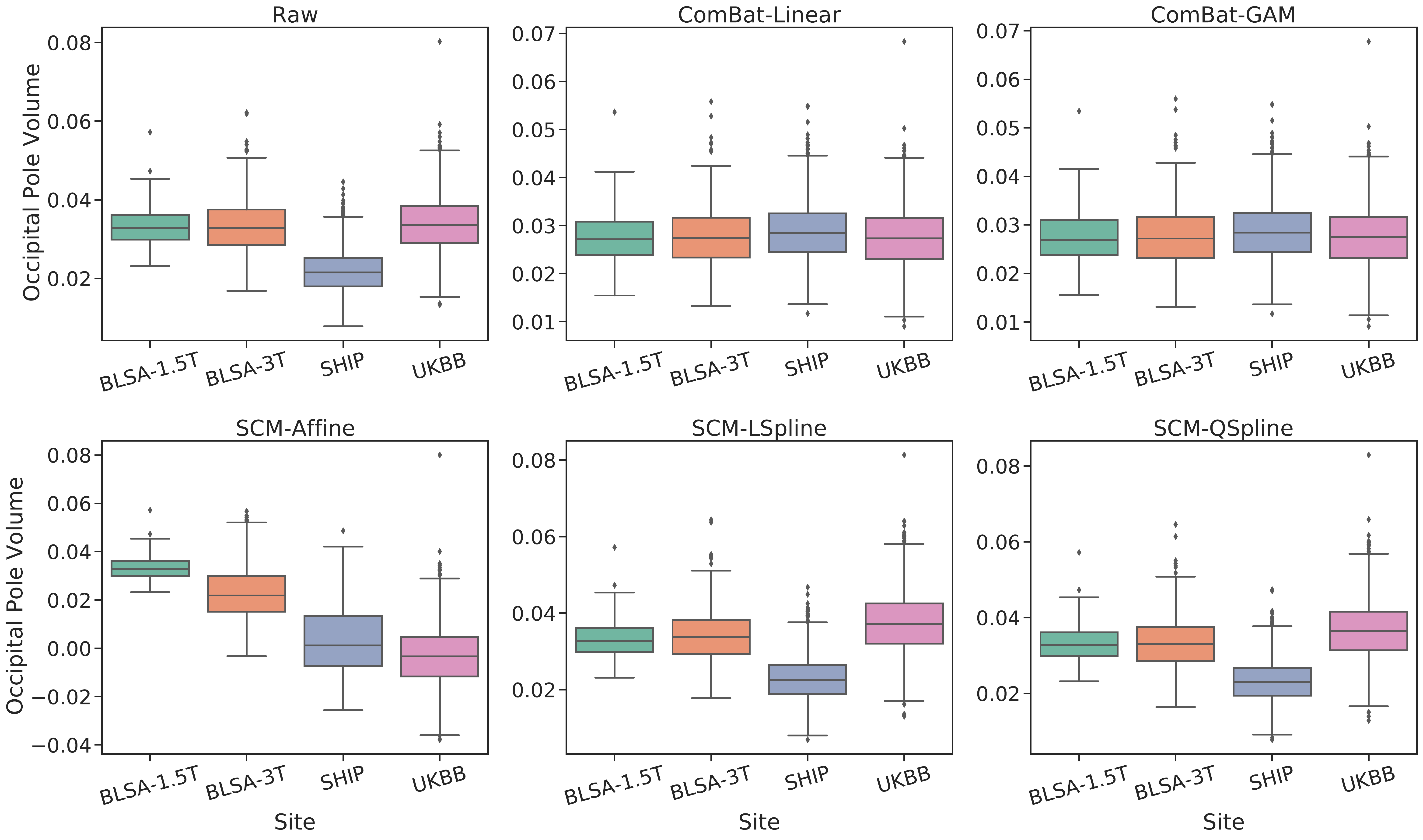}
\caption{Occipital Pole}
\end{subfigure}
\begin{subfigure}[t]{\linewidth}
\centering
\includegraphics[width=\linewidth]{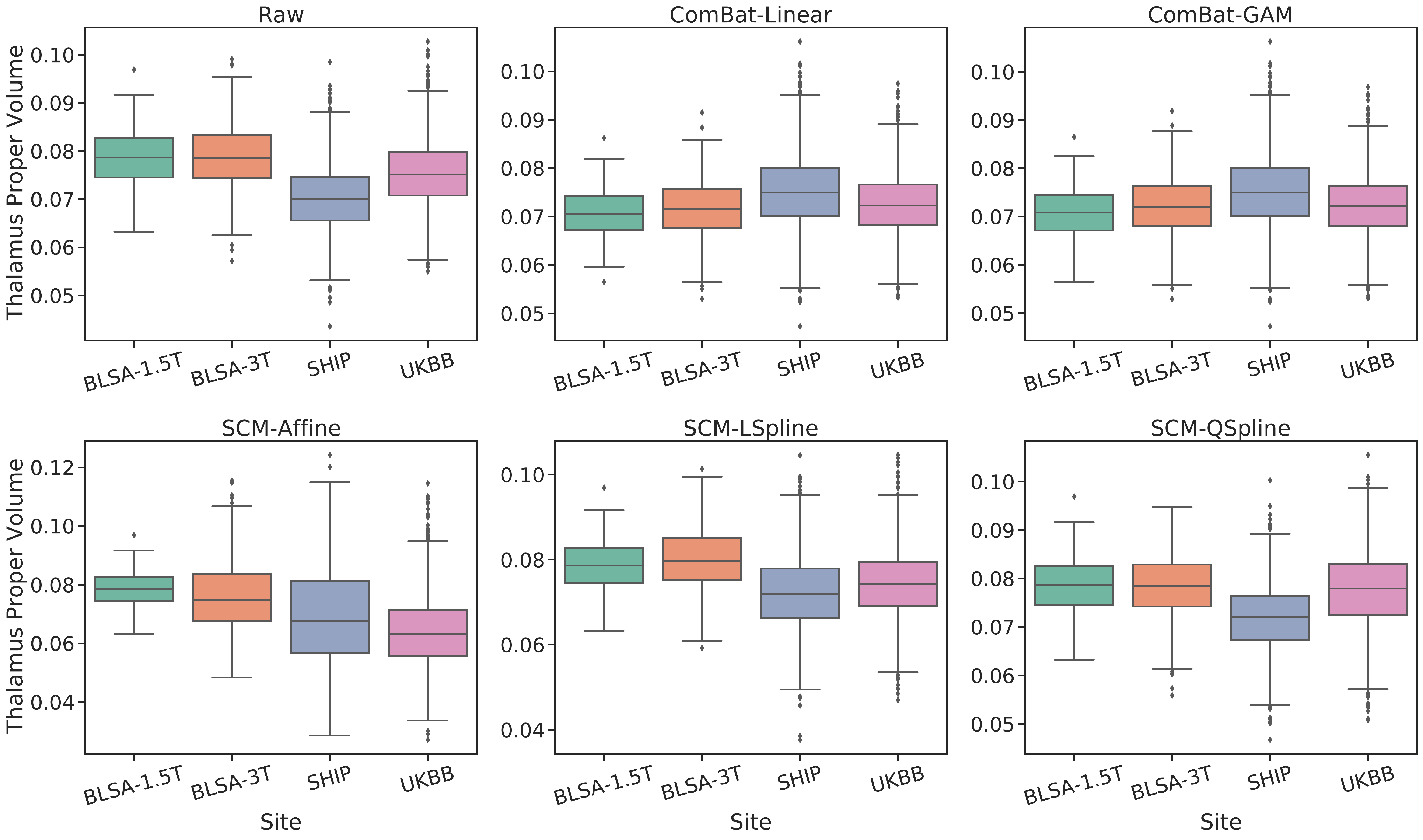}
\caption{Thalamus Proper (Right)}
\end{subfigure}
\caption{Continued comparison of normalized feature distributions cross-site in iSTAGING consortium before and after apply the ComBat methods (ComBat-Linear and ComBat-GAM) and the proposed methods (SCM-Affine, SCM-LSpline, and SCM-QSpline).
The distributions of the features harmonized by ComBat methods are aligned cross-site, whereas those harmonized by our proposed method (Q-Spline) are unchanged compared to the raw features.
We preserve the unknown cofounders (subject-specific information due to biological variability, such as race, gene, and pathology AD/CN) instead of removing them as site-effects, which is beneficial for downstream analysis, such as AD diagnosis.}
\label{fig:occcipital_pole}
\end{figure}

\end{document}